\documentclass{article} 
\usepackage{nips13submit_e,times}
\usepackage{hyperref}
\usepackage{cite}
\usepackage{graphicx}
\usepackage{fancyvrb}
\usepackage{floatrow}
\newfloatcommand{capbtabbox}{table}[][\FBwidth]
\usepackage{blindtext}

\title{Efficient Online Bootstrapping for Large Scale Learning}

\author{
Zhen Qin \\
University of California, Riverside \\
\texttt{zqin001@cs.ucr.edu} \\
\And
Vaclav Petricek \\
eHarmony Inc \\
\texttt{vpetricek@eharmony.com} \\
\AND
Nikos Karampatziakis \\
Microsoft Research \\
\texttt{nikosk@microsoft.com} \\
\And
Lihong Li \\
Microsoft Research \\
\texttt{lihongli@microsoft.com} \\
\And
John Langford \\
Microsoft Research \\
\texttt{jcl@microsoft.com} \\
}

\newcommand{\figref}[1]{Fig.~\ref{#1}}
\newcommand{\tblref}[1]{Tbl.~\ref{#1}}

\nipsfinalcopy 

\begin{document}

\maketitle

\begin{abstract}
Bootstrapping is a useful technique for estimating the uncertainty of a predictor,
for example, confidence intervals for prediction. It is typically used on small
to moderate sized datasets, due to its high computation cost. This work describes a
highly scalable online bootstrapping strategy, implemented inside Vowpal
Wabbit, that is several times faster than traditional strategies. Our experiments
indicate that, in addition to providing a black box-like method for estimating
uncertainty, our implementation of online bootstrapping may also help to train
models with better prediction performance due to model averaging.
\end{abstract}

\section{Introduction}

Bootstrapping is a very common method for sample statistics estimation. It
generates N distinct datasets from the original training data by sampling
examples with replacement; each resampled set is used to train one separate
model. However, instantiating the individual bootstrapped samples is costly, both in
terms of storage and processing. A naive implementation would require N
times the original running time for training only, plus the resampling overhead.
This makes bootstrapping formidable for large-scale learning tasks. 

Vowpal Wabbit\footnote{\url{https://github.com/JohnLangford/vowpal_wabbit}}
(VW) \cite{vw} is a very fast open-source implementation of an online
out-of-core learner. Among the many efficient tricks within, it allocates a
fixed (user-specifiable) memory size for learner representation, implements a hashing
trick that hashes feature names to numeric indexes, and executes parallel threads
for example parsing and learning. VW is able to learn a tera-feature
($10^{12}$) dataset on 1000 nodes in one hour
\cite{DBLP:journals/corr/abs-1110-4198}.

In this work, we extend an online version of bootstrapping for examples with
unit weights proposed by Oza and Russell\cite{Oza01onlinebagging} to arbitrary
positive real-valued weights, taking advantage of the good support of handling varying weights
 in VW \cite{DBLP:conf/uai/KarampatziakisL11}. We provide an efficient implementation of this
algorithm that works as a reduction, and therefore may be used with binary and
multiclass classifiers, regressors, as well as contextual bandit learners. Our
memory efficient strategy scales well to large-scale data. All of
our code is a part of the open-source Vowpal Wabbit project \cite{vw}. 

\section{Background on Online Bootstrapping}

\begin{figure}[ht]
\begin{center}
\includegraphics[height=2.2in]{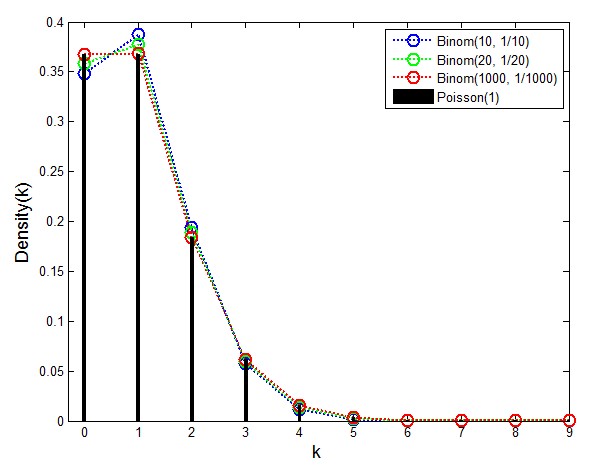}
\end{center}
\caption[Caption for LOF]{Binomial converges to Poisson for moderate sample sizes.}
\label{fig:bvp}
\end{figure}

Online bootstrapping via sampling from Poisson distribution was first proposed
in \cite{Oza01onlinebagging}. This is a very effective online approximation to
batch bootstrapping, leveraging the following argument: Bootstrapping a
dataset $D$ with $n$ examples means sampling n examples from $D$ with
replacement. Each example $i$ will appear $Z_i$ times in the bootstrapped
sample where $Z_i$ is a random variable. In the case of all examples with unit
weight, $Z_i$ is distributed as a $Binom(n,1/n)$, because during resampling the
$i$-th example will have $n$ chances to be picked, each with probability $1/n$.
This $Binom(n,1/n)$ distribution converges to a Poisson distribution with rate
1, even for modest $n$ (see \figref{fig:bvp}). Poisson distribution is much
easier to sample from, making it particularly suitable for large-scale
learning.

Park et al \cite{exact} proposed reservoir sampling with replacement for
sampling streaming data - a technique that could be used to implement
bootstrapping without approximation. However reservoir sampling is more complex
and expensive. Kleiner et al \cite{Kleiner:2012tm} propose a different bootstrap
approximation which divides the large dataset into many little and possibly
non-overlapping subsamples, but each set of subsamples are still processed in a 
batch manner, thus it is not applicable for typical online settings.

\section{Efficient Online Bootstrapping}

Sampling importance weights from a Poisson distribution allows us to implement
an efficient online approximation to bootstrapping. \figref{fig:alg} shows the
basic algorithm. 

\begin{figure}[ht]
\begin{center}
\begin{Verbatim}[commandchars=\\\{\},codes={\catcode`$=3\catcode`_=8}]
Input: example E with importance weight W, 
       user-specified number of bootstrapping rounds N

    Training:               Prediction:
    for i = 1..N            for i = 1..N
    do                      do
      Z ~ Poisson(1) * W      $p_i$ = predict(E, i)
      learn(E, Z, i)        done
    done                    return majority(p) // or mean(p) 
\end{Verbatim}
\end{center}
\vspace{-6pt}
\caption{Algorithm: Efficient Online Bootstrapping}
\label{fig:alg}
\end{figure}

We implemented online bootstrapping as a top-level reduction in VW (see
\figref{fig:alg}). This way bootstrapping can be combined with any other
algorithm implementing \texttt{learn()}. Parameter $i$ is passed to the base learner
to indicate an offset for storing feature weight of the current bootstrapping
submodel. This architecture has three benefits: i) It keeps bootstrapping code
separate from the learning code, capitalizing on any improvements in the
base learning algorithm, ii) Weights for the same feature inside different
bootstraps can be co-located in memory, which keeps memory access local and
maximizes cache hits, and iii) Each example needs to be parsed only once for all
bootstrapped submodels. Only the importance weight of example is modified by
drawing repeatedly from a Poisson distribution.  This greatly reduces example
parsing overhead.  

There are two alternatives to implement online bootstrapping for
non-unitary weights: i) sample the new importance weight directly from
$Poisson(W)$ or ii) sample from a $Poisson(1)$ and multiply it with $W$. Option
i) suffers when $W \ll 1$ as it almost always rejects in this case (most weights
are zero).  Option ii) is preferable and can be implemented very efficiently by
a lookup table of the $Poisson(1)$ probabilities.\footnote{Only 20 entries are
needed before the probability drops below machine precision}.


During prediction (testing), each example is again parsed only once and fed into N
learners online. Besides estimating statistics from these N predictions, user
can specify different ensemble methods to get one final prediction. The current
implementation supports mean (for regression) and majority voting (for
classification). Implementation of other statistics including quantiles is
straightforward.

\section{Experiments}
\begin{figure*}[h]
	\centering
\begin{tabular}{c@{\hspace{2pt}}c}
	\includegraphics[height=1.8in]{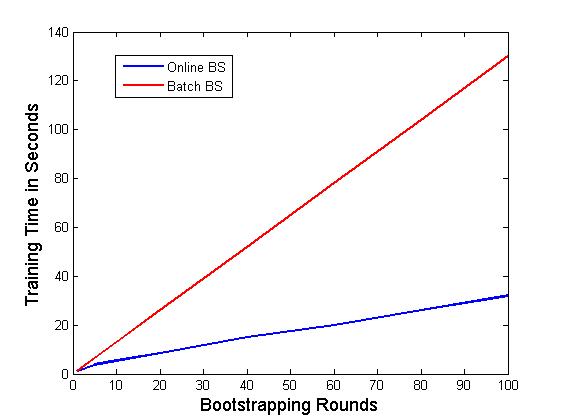} &
  \includegraphics[height=1.8in]{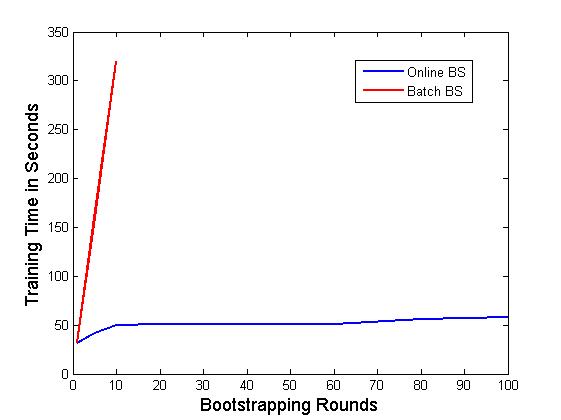} \\
  75K Dataset & RCV1 Training \\
\end{tabular}
\vspace{-6pt}
  \caption{Speed comparisons on two datasets. Lower bound is calculated (some numbers are not shown since they are too big) for Batch BS as pure training time, excluding time to generate samples with repetition. }
  \label{fig:time}
\end{figure*}

We show the efficiency of our online bootstrapping strategy, as well as how it
helps to improve prediction accuracy. All experiments are conducted on a single
desktop. We first show speed comparison on two datasets. The 75K dataset
contains 74746 examples and 3000 features per example on average. For this
dataset, we run 20 online passes to mitigate setting up overhead. The RCV1
\cite{Lewis:2004:RNB:1005332.1005345} training dataset contains 781265 examples
and 80 features per example on average. We only run single pass for this
dataset. Running time for batch bootstrapping is estimated as $t \times n$
where n is number of bootstrap samples and $t$ is time for $n=1$. We believe
this estimation is optimistic, as it is not even clear how to do batch
resampling on large datasets. We show results in \figref{fig:time}. It is clear
there is more performance gain for the RCV1 training dataset.  Also the running
time does not differ too much with different number of bootstrapping rounds
for the RCV1 training dataset. These can be explained as a lot of 
computation power is spent on example parsing, while one benefit
of our approach is to avoid repetitive example parsing.
Thus our strategy is particularly helpful for a dataset with many examples.


\begin{figure}[h]
\begin{floatrow}
\ffigbox{%
  \includegraphics[height=1.9in]{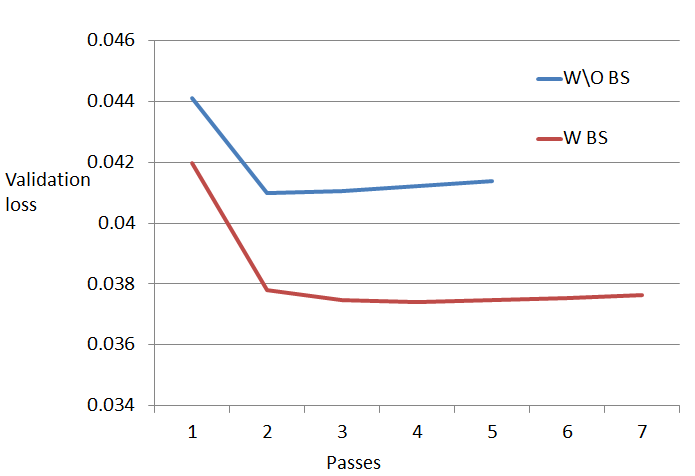}
}{%
  \caption{Bootstrapping helps predictive performance compared to a single model.}\label{curve}%
}
\capbtabbox{%
  \begin{tabular}{cc} \hline
  Method & Error Rate \\ \hline
  Base Learner (BL) &  6.01\% \\
  BL + Online BS (N=20) & 5.37\% \\
  Tuned Learner (TL) & 4.64\% \\
  TL + Online BS (N=4) & 4.58\% \\ \hline
  \medskip
  \medskip
  \medskip
  \end{tabular}
}{%
  \caption{Testing performance}\label{testing}%
}
\end{floatrow}
\end{figure}

Next we show that online bootstrapping may improve predictions. We train on the
RCV1 training dataset and report the online holdout validation loss\footnote{A
feature available in VW. This loss is calculated online on a consistent set of
examples across passes that are used for model evaluation but not for model
updating} at the end of each pass. Experiments are conducted with 20 bootstraps
and $2^{24}$ bits used for learner representation. We use square loss for all 
experiments. \figref{curve} shows that with
bootstrapping the online validation loss is indeed lower. We save the best model
according to validation loss and test generalization on a separate test set
containing 23149 examples. In \tblref{testing}, we measure classification accuracy
and can see that the model trained with online bootstrapping performs better. We further add
bootstrapping to a well tuned learner \footnote{Single pass learning with options -b 23
-l 0.25 -{}-ngram 2 -{}-skips 4} and observe that online bootstrapping can improve performance
upon such a competitive baseline.

Bootstrapping learns more parameters than a single model and therefore for a
fixed model size increases the chance of hashing collisions. This effectively
leads to learning N simpler models versus a single more complex model. We ran
three tests on the RCV1 dataset to investigate this trade-off: i) single model
(default), ii) single model with extra quadratic features and iii) bootstrapped
model with extra quadratics.  We ran multiple passes and saved the best model
based on holdout loss. The holdout loss results are summarized in
\tblref{bsandq}. We can see that on this dataset even with larger number of
features, bootstrapping helps to improve model performance.

\begin{table}[h!]
\caption{Bootstrapping may help even in the presence of some collisions.}
\label{bsandq}
\begin{center}
\begin{tabular}{lcrr}
\\ \hline 
               & squared loss & number of features & model \\
Model          & of best model & per example & size\\ \hline
default        &    0.0409 & 80 on average& $2^{24}$\\
+quadratic     &    0.0402 & 3-40K& $2^{24}$\\
+quadratic + BS &    0.0340 & 60-800K& $2^{24}$\\
\end{tabular}
\end{center}
\end{table}

\section{Conclusions}

In this work we show a highly effective and efficient online bootstrapping
strategy that we implemented in the open-source Vowpal Wabbit online learning
package. It is fast, feasible for large scale datasets, improves predictions of
the resulting model, and provides a blackbox-like way to obtain uncertainty
estimates that works with a variety of existing learning algorithms.



\pagebreak

\subsubsection*{References}
\vspace{-6pt}
\renewcommand{\refname}{}
\bibliography{onlinebs}{}

\begin{thebibliography}{1}

\bibitem{DBLP:journals/corr/abs-1110-4198}
Alekh Agarwal, Olivier Chapelle, Miroslav Dud\'{\i}k, and John Langford.
\newblock A reliable effective terascale linear learning system.
\newblock {\em CoRR}, abs/1110.4198, 2011.

\bibitem{DBLP:conf/uai/KarampatziakisL11}
Nikos Karampatziakis and John Langford.
\newblock Online importance weight aware updates.
\newblock In {\em UAI}, pages 392--399, 2011.

\bibitem{Kleiner:2012tm}
Ariel Kleiner, Ameet Talwalkar, Purnamrita Sarkar, and Michael~I. Jordan.
\newblock {The Big Data Bootstrap}.
\newblock In {\em ICML}, 2012.

\bibitem{vw}
John Langford, Lihong Li, and Alexander Strehl.
\newblock Vowpal wabbit open source project.
\newblock In {\em Technical Report, Yahoo!}, 2007.

\bibitem{Lewis:2004:RNB:1005332.1005345}
David~D. Lewis, Yiming Yang, Tony~G. Rose, and Fan Li.
\newblock Rcv1: A new benchmark collection for text categorization research.
\newblock {\em J. Mach. Learn. Res.}, 5:361--397, December 2004.

\bibitem{Oza01onlinebagging}
Nikunj~C. Oza and Stuart Russell.
\newblock Online bagging and boosting.
\newblock In {\em In Artificial Intelligence and Statistics 2001}, pages
  105--112. Morgan Kaufmann, 2001.

\bibitem{exact}
Byung-Hoon Park, George Ostrouchov, and Nagiza~F. Samatova.
\newblock Sampling streaming data with replacement.
\newblock {\em Comput. Stat. Data Anal.}, 52(2):750--762, October 2007.

\end{thebibliography}
\bibliographystyle{plain}
\end{document}